\documentclass[11pt,a4paper]{article}
\usepackage[hyperref]{emnlp2021}
\usepackage{amsthm}

\usepackage{times}
\usepackage{latexsym}
\usepackage{amsmath}
\usepackage{xfrac}
\usepackage{amsfonts}
\usepackage{amssymb}
\usepackage{mathtools}
\usepackage{url}
\usepackage{microtype}
\usepackage{inconsolata}
\usepackage{multirow}
\usepackage{booktabs}
\usepackage{adjustbox}
\usepackage{tikz}
\usepackage{algorithm}
\usepackage{caption}
\usepackage{algpseudocode}
\usepackage{subcaption}
\usepackage{color}
\usepackage[safe]{tipa}
\usetikzlibrary{positioning,chains,fit,shapes,calc}

\usepackage{todonotes}
\newcommand{\matching}{{\mymacro{\boldsymbol{m}}}}
\newcommand{\matchings}{{\mymacro{\mathcal{M}}}}
\newcommand{\alignment}{{\mymacro{\boldsymbol{a}}}}
\newcommand{\alignments}{\mymacro{{\mathcal{A}}}}
\newcommand{\ntrg}{{\mymacro{n_{\textit{trg}}}}}
\newcommand{\nsrc}{{\mymacro{n_{\textit{src}}}}}
\newcommand{\bigO}{{\mymacro{\mathcal{O}}}}

\usepackage{thm-restate}

\definecolor{myblue}{RGB}{80,80,160}
\definecolor{mygreen}{RGB}{80,160,80}
\definecolor{myred}{RGB}{160,80,80}

\usepackage{xcolor}
\definecolor{nice-red}{HTML}{E41A1C}
\definecolor{nice-orange}{HTML}{FF7F00}
\definecolor{nice-yellow}{HTML}{FFC020}
\definecolor{nice-green}{HTML}{4DAF4A}
\definecolor{nice-blue}{HTML}{377EB8}
\definecolor{nice-purple}{HTML}{984EA3}

\DeclareMathOperator*{\argmax}{{\mymacro{ argmax}}}

\newcommand{\vtheta}{\boldsymbol{\theta}}
\newcommand{\ptheta}{{\mymacro{p_{\vtheta}}}}
\newcommand{\R}{\mathbb{R}}
\newcommand{\mymacro}[1]{#1}
\newcommand{\ltrg}{{\mymacro{\ell_\textit{trg}}}}
\newcommand{\lsrc}{{\mymacro{\ell_\textit{src}}}}
\newcommand{\Vtrg}{{\mymacro{V_\textit{trg}}}}
\newcommand{\Vsrc}{{\mymacro{V_\textit{src}}}}
\newcommand{\vtrg}{{\mymacro{v_\textit{trg}}}}
\newcommand{\vsrc}{{\mymacro{v_\textit{src}}}}
\newcommand{\utrg}{{\mymacro{\boldsymbol{u}_\textit{trg}}}}
\newcommand{\usrc}{{\mymacro{\boldsymbol{u}_\textit{src}}}}
\newcommand{\word}[1]{{\mymacro{\textit{#1}}}}
\newcommand{\normal}{{\mymacro{\mathcal{N}}}}

\makeatletter
\def\moverlay{\mathpalette\mov@rlay}
\def\mov@rlay#1#2{\leavevmode\vtop{%
   \baselineskip\z@skip \lineskiplimit-\maxdimen
   \ialign{\hfil$\m@th#1##$\hfil\cr#2\crcr}}}
\newcommand{\charfusion}[3][\mathord]{
    #1{\ifx#1\mathop\vphantom{#2}\fi
        \mathpalette\mov@rlay{#2\cr#3}
      }
    \ifx#1\mathop\expandafter\displaylimits\fi}
\makeatother

\newcommand{\cupdot}{\charfusion[\mathbin]{\cup}{\cdot}}

\usepackage{cleveref}
\crefname{section}{\S}{\S\S}
\crefname{table}{Tab.}{Tabs.}
\crefname{figure}{Fig.}{Figs.}
\crefname{algorithm}{Algorithm}{Algorithms}
\crefname{equation}{Eq.}{Eqs.}
\crefname{example}{Example}{Examples}
\crefname{fact}{Fact}{Facts}
\crefname{appendix}{Appendix}{Appendices}
\crefname{theorem}{Theorem}{Theorems}
\crefname{reTheorem}{Theorem}{Theorems}
\crefname{reProposition}{Proposition}{Propositions}
\crefname{aquestion}{Question}{Questions}
\crefname{assumption}{Assumption}{Assumptions}
\crefname{lemma}{Lemma}{Lemmas}
\crefname{reLemma}{Lemma}{Lemmas}
\crefname{proposition}{Proposition}{Propositions}
\crefname{chapter}{Chapter}{Chapters}
\crefname{line}{line}{lines}
\crefname{principle}{Principle}{Principles}
\crefname{definition}{Definition}{Definitions}
\crefname{corollary}{Corollary}{Corollaries}
\crefname{Exercise}{Exercise}{Exercises}
\crefformat{section}{\S#2#1#3}

\newcommand{\coloneqq}{\mathrel{\stackrel{\textnormal{\tiny def}}{=}}}

\title{A Discriminative Latent-Variable Model for Bilingual Lexicon Induction}

\author  
  {
	\begin{tabular}{cccc}
	Sebastian Ruder\raise1.0ex\hbox{\normalfont\normalsize \textschwa}\raise1.0ex\hbox{\normalfont \normalsize,\textipa{H}}\footnotemark[1] & Ryan Cotterell\raise1.0ex\hbox{\normalfont\normalsize \textipa{S}}\raise1.0ex\hbox{\normalfont\normalsize ,\textipa{P}}\footnotemark[1] & Yova Kementchedjhieva\raise1.0ex\hbox{\normalfont\normalsize \textipa{Z}} & Anders S{\o}gaard\raise1.0ex\hbox{\normalfont\normalsize \textipa{Z}}
	\end{tabular}
	\\
    \raise1.0ex\hbox{\normalfont\normalsize \textschwa}Insight Research Centre, National University of Ireland, Galway, Ireland\\
    \raise1.0ex\hbox{\normalfont \normalsize \textipa{H}}Aylien Ltd., Dublin, Ireland\\
    \raise1.0ex\hbox{\normalfont\normalsize \textipa{S}}The Computer Laboratory, University of Cambridge, Cambridge, UK\\
    \raise1.0ex\hbox{\normalfont\normalsize \textipa{P}}Department of Computer Science, Johns Hopkins University, Baltimore, USA \\
     \raise1.0ex\hbox{\normalfont\normalsize \textipa{Z}}Department of Computer Science, University of Copenhagen, Copenhagen, Denmark\\
	{\href{mailto:sebastian@ruder.io}{\texttt{sebastian@ruder.io}} \quad \href{mailto:ryan.cotterell@jhu.edu}{\texttt{ryan.cotterell@jhu.edu}} \quad \{\href{mailto:yova@di.ku.dk}{\texttt{yova}}, \href{mailto:soegaard@di.ku.dk}{\texttt{soegaard}}\}\texttt{@di.ku.dk}}
}

\date{}

\begin{document}
\maketitle

\newenvironment{starfootnotes}
  {\par\edef\savedfootnotenumber{\number\value{footnote}}
   \renewcommand{\thefootnote}{$*$} 
   \setcounter{footnote}{0}}
  {\par\setcounter{footnote}{\savedfootnotenumber}}

\begin{starfootnotes}
\footnotetext{The first two authors contributed equally.}
\end{starfootnotes}

\begin{abstract}
We introduce a novel discriminative latent-variable model for bilingual lexicon induction. 
Our model combines the bipartite matching dictionary prior of \citet{Haghighi2008} with a representation-based approach \citep{Artetxe2017}. 
To train the model, we derive an efficient Viterbi EM algorithm. 
We provide empirical results on six language pairs under two metrics and show that the prior improves the induced bilingual lexicons.
We also demonstrate how previous work may be viewed as a similarly fashioned latent-variable model, albeit with a different prior.\footnote{\url{https://github.com/sebastianruder/latent-variable-vecmap}}
\end{abstract}

\section{Introduction}
Is there a more fundamental bilingual linguistic resource than a
dictionary?  
The task of bilingual lexicon induction seeks to create a
dictionary in a data-driven manner directly from monolingual corpora
in the respective languages and, perhaps, a small seed set of
translations. 
From a practical point of view, bilingual dictionaries have found uses in a myriad of NLP tasks ranging from machine translation \citep{klementiev-EtAl:2012:EACL2012} to cross-lingual named entity
recognition \citep{mayhew-tsai-roth:2017:EMNLP2017}.
In this work, we offer a probabilistic twist on the task, developing 
a novel discriminative latent-variable model that outperforms previous work.

Our proposed model is a bridge between current state-of-the-art
methods in bilingual lexicon induction that take advantage of word
representations, e.g., the representations induced by \citeposs{mikolov2013efficient}
skip-gram objective, and older ideas in the literature that build an explicit probabilistic
model for the task. 
In contrast to previous work, our model is a discriminative probability model, inspired by
\citet{irvine-callisonburch:2013:NAACL-HLT}, but infused with the bipartite matching dictionary prior of \citet{Haghighi2008}.
However, similar to more recent approaches \citep{Artetxe2017}, our model operates directly over
word representations, inducing a joint cross-lingual representation space, and scales to large vocabulary sizes. 
To train our model, we derive a generalized expectation maximization algorithm
\cite[EM; ][]{nh-nveajis-98} and employ an efficient combinatorial algorithm to perform the bipartite matching.\looseness=-1

Empirically, we experiment on three high-resource and three extremely low-resource language pairs. 
We evaluate intrinsically, comparing the quality of the induced bilingual dictionary, as well as analyzing the resulting bilingual word representations themselves. 
The latent-variable model yields gains over several previous approaches across language pairs. 
It also enables us to make implicit modeling assumptions explicit. 
To this end, we provide a reinterpretation of \citet{Artetxe2017} as a latent-variable model in the style of IBM Model 1 \citep{Brown:1993:MSM:972470.972474}, which allows a clean side-by-side analytical comparison between our method and previous work. 
Viewed in this light, the difference between our approach and
\citeposs{Artetxe2017} approach, the strongest baseline, is whether one-to-one alignments or one-to-many alignments are admitted between the words of the languages' respective lexicons.
We conclude our prior over one-to-one alignments is primarily responsible for the improvements over \citeposs{Artetxe2017} model, which allows more permissive alignments.\looseness=-1

\section{Background}\label{sec:background}
Bilingual lexicon induction\footnote{In this paper, we use bilingual lexicon and (bilingual) dictionary synonymously. On the other hand, unqualified use of lexicon refers to a word list in a single language.} is the task of finding word-level
translations between the lexicons of two languages. For instance, the
German word \word{Hund} and the English word \word{dog} are roughly
semantically equivalent, so the pair \word{Hund}--\word{dog} should be
an entry in a German--English bilingual lexicon. The task itself comes
in a variety of flavors. 
In this paper, we consider a version of the task that only relies on \emph{monolingual} corpora in the tradition of \citet{rapp:1995:ACL} and \citet{W95-0114}.  
In other words, the goal is to produce a bilingual lexicon primarily from \emph{unannotated} raw text in each of the respective languages.
Importantly, we avoid reliance on bitext, i.e., corpora with parallel sentences that are known translations of each other, e.g., EuroParl \citep{koehn2005europarl}. 
The bitext assumption is quite common in the literature; see \citet[Table 2]{DBLP:journals/corr/Ruder17} for a survey.  
Additionally, we will assume the existence of a small seed set of word-level translations obtained from a dictionary; we also experiment with seed sets obtained from heuristics that do not rely on the existence of linguistic resources.

\subsection{Graph-Theoretic Formulation}\label{sec:graph-theory}
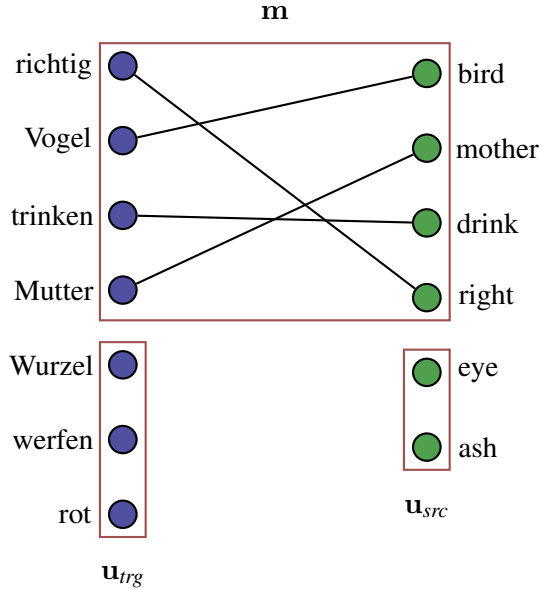
\begin{figure}
\begin{tikzpicture}[thick,
  every node/.style={draw,circle},
  fsnode/.style={fill=myblue},
  ssnode/.style={fill=mygreen},
  %every fit/.style={ellipse,draw,inner sep=-1pt,text width=3cm},
  %-,shorten >= 2pt,shorten <= 2pt,
  msnode/.style={rectangle,inner sep=3pt,text width=-10cm,draw=myred,thick},
]

% the vertices of U
\begin{scope}[start chain=going below,node distance=6mm]
\foreach \i in {richtig,Vogel,trinken,Mutter,Wurzel,werfen,rot}
  \node[fsnode,on chain] (f\i) [label=left: \i] {};
\end{scope}

% the vertices of V
\begin{scope}[xshift=4cm,yshift=-0.1cm,start chain=going below,node distance=6mm]
\foreach \i in {bird,mother,drink,right,eye,ash}
  \node[ssnode,on chain] (s\i) [label=right: \i] {};
\end{scope}

% the set U
\node [msnode,fit=(frichtig) (sright),label=above:$\matching$] {};
\node [msnode,fit=(fWurzel) (frot),label=below:$\utrg$] {};
\node [msnode,fit=(seye) (sash),label=below:$\usrc$] {};
% the set V
%\node [mygreen,fit=(s6) (s9),label=above:$V$] {};

% the edges
\draw (frichtig) -- (sright);
\draw (fVogel) -- (sbird);
\draw (fMutter) -- (smother);
\draw (ftrinken) -- (sdrink);
%\draw (s6) -- (f2);
%\draw (f2) -- (s7);
%\draw (s7) -- (f3);
%\draw (s8) -- (f3);
%\draw (f3) -- (s9);
%\draw (s9) -- (f5);
%\draw (f5) -- (s6);
\end{tikzpicture}
\caption{Partial lexicons of German and English shown as a bipartite graph. German is the target
  language and English is the source language. The $\ntrg=7$ German words are shown in blue and the $\nsrc=6$ English words are shown in green. A bipartite matching $\matching$ between the two sets of vertices is also depicted. The German 
  nodes in $\utrg$ are unmatched.}
\label{fig:graph}
\vspace{-10pt}
\end{figure}

To ease the later exposition, we will formulate the task of bilingual lexicon induction graph-theoretically. 
Let $\lsrc$ denote the source language and $\ltrg$ the target language. 
Suppose the source language $\lsrc$ has $\nsrc$ word types in its lexicon $\Vsrc$ and $\ltrg$ has $\ntrg$ word types in its lexicon~$\Vtrg$.
We will write $\vsrc(i)$ for the $i^\text{th}$ word type in $\lsrc$ and $\vtrg(i)$ for the $i^\text{th}$ word type in $\ltrg$. 
We will often write $i$ for $\vtrg(i)$ and $j$ for $\vsrc(j)$ for brevity.
Now consider the bipartite set of vertices $\Vtrg\cupdot\Vsrc$. 
In these terms, a bilingual lexicon is just a \textbf{bipartite graph} $G = (\Vtrg\cupdot\Vsrc, E)$ and, thus, the
task of bilingual lexicon induction is a combinatorial problem: the
search for a good edge set $E \subseteq \Vtrg \times \Vsrc$. 
We depict such a bipartite graph in \cref{fig:graph}. 
In \cref{sec:model}, we will operationalize the notion of goodness by assigning a weight $w_{ij}$ to each possible edge between $\Vtrg$ and $\Vsrc$.\looseness=-1

When the edge set $E$ takes the form of a \textbf{bipartite matching}, we will
denote it as $\matching$.
In general, we will be interested in partial bipartite matchings, i.e., bipartite matchings where some vertices may have \emph{no} incident edges. 
We will write $\matchings$ for the set of all partial matchings on the bipartite graph~$G$. 
The set of vertices in $\Vtrg$ (respectively, $\Vsrc$) with no incident edges
will be termed $\utrg$ (respectively, $\usrc$). 
Note that for any matching $\matching$, we have the identity $\utrg = \Vtrg \setminus \{ i : (i, j) \in \matching\}$.\looseness=-1

\subsection{Word Representations}\label{sec:representations}
Word representations will also play a key role in our
model. For the remainder of the paper,
we will assume we have access to $d$-dimensional
real-valued representations for each language's lexicon---for example, those provided
by a standard model such as skip-gram \citep{mikolov2013efficient}.
Notationally, we define the real matrices~$S \in
\R^{d \times \nsrc}$ and~$T \in
\R^{d \times \ntrg}$. 
Note that in this formulation $s_i
\in \R^d$, the $i^\text{th}$ column of~$S$, is the word
representation corresponding to the word $\vsrc(i)$. 
Likewise, note $t_i \in \R^d$, the $i^\text{th}$ column of $T$, is the word representation corresponding to the word $\vtrg(i)$.

\section{A Latent-Variable Model}\label{sec:model}
The primary contribution of this paper is a novel latent-variable
model for bilingual lexicon induction.  The latent variable will be
the edge set $E$, as discussed in \cref{sec:graph-theory}. 
We define the following density
\begin{equation}
  \ptheta(T \mid S) \coloneqq \sum_{\matching \in \matchings} \ptheta(T \mid S, \matching) p(\matching)
\end{equation}
over representations of the target language given the representations of the source language.
Recall from \cref{sec:background}, $\matchings$ is the set of all bipartite matchings on the graph $G$ and $\matching \in \matchings$ is an individual matching. 
Note that,
then, $p(\matching)$ is a \emph{distribution} over all partial bipartite matchings on $G$
such as the matching shown in \cref{fig:graph}. 
We will take $p(\matching)$ to be fixed as the uniform distribution for the remainder of the exposition, though more complicated distributions could be learned. 
We further define the density
\begin{equation}
  \ptheta(T \mid S, \matching) \coloneqq \prod_{(i, j) \in \matching}  p(t_i \mid s_j) \prod_{i \in \utrg} p(t_i) \label{eq:model}
\end{equation}
where we write $(i, j) \in \matching$ to denote an edge in the bipartite matching. 
Furthermore, for notational simplicity, we have dropped the dependence of $\utrg$ on $\matching$---recall $\utrg = \Vtrg \setminus \{ i : (i, j) \in \matching\}$. 
Next, we define the two densities present in \Cref{eq:model}:
\begin{subequations}
\begin{align}
  \ptheta(t \mid s) &\coloneqq \normal\left(\Omega\, s, I \right) \label{eq:conditional}\\
  &\propto \exp -\sfrac{1}{2}||t - \Omega \,s||_2^2 \\
  \ptheta(t) &\coloneqq  \normal\left(\mu, I\right) \label{eq:non-conditional} \\
  &\propto \exp -\sfrac{1}{2}||t - \mu||_2^2 
\end{align}
\end{subequations}
where $\Omega \in \R^{d \times d}$ is a real orthogonal matrix of parameters to be learned. 
We define the model's parameters, to be optimized, as $\vtheta = \left(\Omega, \mu\right)$, which justifies our use of the subscript $\vtheta$ in $\ptheta$.

Now, given a fixed matching $\matching$, we may create submatrices $S_{\matching} \in \R^{d \times |\matching|}$ and $T_{\matching} \in \R^{d \times |\matching|}$ such that the columns
correspond to word vectors of matched vertices, i.e., translations under the bipartite matching $\matching$.  
Now, after some algebra, we can rewrite $\prod_{(i, j) \in \matching} p(t_i \mid s_j)$ in matrix notation:
\begin{subequations}
\begin{align}
  \ptheta(T_{\matching} \mid\,& S_{\matching}, \matching) = \prod_{(i, j) \in \matching} p(t_i \mid s_j) \\ 
  &\propto \prod_{(i, j) \in \matching} \exp -\sfrac{1}{2}||t_i - \Omega\, s_j||_2^2 \\ 
  & = \exp \sum_{(i ,j) \in \matching} -\sfrac{1}{2}||t_i - \Omega\, s_j ||_2^2 \\ 
  & = \exp -\sfrac{1}{2}||T_{\matching} - \Omega \, S_{\matching}||_{\mathrm{F}}^2  \label{sec:after-algebra}
\end{align}
\end{subequations}
The result of this derivation, \cref{sec:after-algebra}, will become
useful during the discussion in \cref{sec:parameter-estimation}.\looseness=-1

\paragraph{Modeling Assumptions.}
In the previous section, we have formulated the induction of a bilingual
lexicon as the search for an edge set $E$, which
we treat as a latent variable in a discriminative probability model.
Specifically, we assume that $E$ is a partial
matching. 
Thus, for every $(i, j) \in \matching$, we have $t_i \sim
\normal\left(\Omega\, s_j, I\right)$, that is, the representation
for $\vtrg(i)$ is assumed to have been drawn from a Gaussian centered around the representation for $\vsrc(j)$, after an orthogonal transformation. 
This gives rise to two modeling assumptions, which we make explicit:
(i) There exists a single source for every word in the target lexicon
\emph{and} that source cannot be used more than once.
(ii) There exists an orthogonal transformation, after which the representation spaces
are more or less equivalent.

Assumption (i) may be true for related languages, but is likely false for morphologically rich languages that have a many-to-many relationship between the words in their respective lexicons. 
Empirically, we ameliorate this problem with a frequency constraint that only considers the top-$n$ most frequent words in both lexicons for matching in \cref{sec:experiments}. 
In addition, we experiment with priors that relax this constraint in \cref{sec:analysis}.
As for assumption (ii), previous work \citep{Xing2015,Artetxe2017} has
achieved some success using an orthogonal transformation.
Recently, however, \citet{Sogaard2018} demonstrated that monolingual representation spaces are not approximately isomorphic and that there is a complex relationship between word form and meaning, which is only inadequately modeled by current approaches, which for example cannot model polysemy. 
Nevertheless, we show that imbuing our model with these assumptions helps empirically in \cref{sec:experiments}.\looseness=-1

\paragraph{Why it Works: The Hubness Problem.}
Why should we expect the bipartite matching prior to help, given
that we know of cases when multiple source words \emph{should match a target word}?
One answer is because the bipartite prior helps us obviate the \textbf{hubness problem},
a common issue in word-representation-based
bilingual lexicon induction \citep{Dinu2015}.
The hubness problem is an intrinsic problem of high-dimensional vector spaces
where certain vectors will be \emph{universal} nearest neighbors, i.e.,
they will be the nearest neighbor to a disproportionate number of
other vectors \citep{journals/jmlr/RadovanovicNI10}. Thus, if we allow
one-to-many alignments, we will find the representations of certain elements of
$\Vsrc$ acting as hubs, i.e., the model will pick them to generate
a disproportionate number of target representations, which reduces the quality of the representation space.
Another explanation for the positive effect of the one-to-one
alignment prior is its connection to the Wasserstein distance and
computational optimal transport \citep{villani2008optimal, Grave2018}.\looseness=-1

\section{Parameter Estimation}\label{sec:parameter-estimation}
We conduct parameter estimation via Viterbi EM. 
We describe first the E-step, then the M-step. 
Viterbi EM estimates the parameters by alternating between the two steps until convergence. 
We give the complete pseudocode in \cref{alg:viterbi}.

\subsection{Viterbi E-Step}\label{sec:e-step}
The E-step asks us to compute the posterior of latent bipartite
matchings $\ptheta(\matching \mid S, T)$ given the representations $S$ and $T$.  
Computation of this distribution, however, is intractable as it would require a \emph{sum}
over all bipartite matchings, which is \#P-hard \citep{valiant1979complexity}. 
However, tricks from combinatorial optimization make it possible to \emph{maximize}
over all bipartite matchings in polynomial time.  
Thus, we fall back on the Viterbi approximation for the E-step  \citep{Brown:1993:MSM:972470.972474,samdani-chang-roth:2012:NAACL-HLT}. 
The Viterbi E-step requires us to solve the following combinatorial  optimization problem
\begin{equation}\label{eq:e-step}
   \matching^\star = \argmax_{\matching \in \matchings} \log \ptheta(\matching \mid S, T)
\end{equation}
In the following Proposition, we give a polynomial-time solution to \Cref{eq:e-step}.
\begin{restatable}{reProposition}{propATwo}\label{prop:matching}
The optimization problem $\argmax_{\matching \in \matchings} \log \ptheta(\matching \mid S, T)$
can be solved in $\bigO((\nsrc+\ntrg)^3)$ time with the Hungarian algorithm \citep{kuhn1955hungarian}.
\end{restatable}

\paragraph{Exploiting Sparsity.}
The solution given in \cref{prop:matching} requires use of the Hungarian algorithm.
Despite a runtime of $\bigO((\nsrc+\ntrg)^3)$, preliminary experimentation found it too slow for practical use with large vocabulary sizes.\footnote{For reference, in \cref{sec:experiments}, we learn bilingual lexicons between representations of 200,000 source and target language words.}
Thus, we consider a sparsification heuristic.
For each source word, most target words, however, are unlikely candidates for alignment. We thus only consider the top-$k$ most similar target words for alignment with every source word. 
After the sparsification, we employ a version of the Jonker--Volgenant algorithm  \citep{jonker1987shortest,volgenant1996linear}, which has been optimized for bipartite matching on sparse graphs.\footnote{After acceptance to EMNLP 2018, Edouard Grave pointed out that Sinkhorn propagation \citep{adams2011ranking,mena2018learning} may have been a computationally more effective way to deal with the latent matchings.}

\paragraph{Relationship to Cosine Distance.}
In \Cref{prop:matching}, we weight the induced bipartite graph with edge weights
\begin{equation}\label{eq:graph-weight}
    w_{ij} \coloneqq \log \ptheta(t_i \mid s_j) - \log \ptheta(t_i)
\end{equation}
on $(\vtrg(i), \vsrc(j))$ where $\vtrg(i)$ is the vertex corresponding to the $i^{\text{th}}$ word in the target language and $\vsrc(j)$ is the $j^{\text{th}}$ word in the source language. 
We give a simple Proposition relating \Cref{eq:graph-weight} to cosine distance.
\begin{restatable}{reProposition}{propAOne}\label{prop:cosine}
Let $\ptheta(t \mid s)$ be defined as in \Cref{eq:conditional}, and let $\ptheta(t)$ be defined as in \Cref{eq:non-conditional}.
Then, if $||t||_2 = 1$ and $||s||_2 = 1$, we have
\begin{equation}
    \log \frac{\ptheta(t \mid s)}{\ptheta(t)} = t^\top\!\left(\Omega\,s - \mu\right) + \sfrac{1}{2}\left(||\mu||_2^2 - 1\right)
\end{equation}
In particular, when $\mu = \mathbf{0}$, this reduces to $\cos(t, \Omega\,s) - \sfrac{1}{2}$.
\end{restatable}
\Cref{prop:cosine} justifies the use of cosine distance as a graph-weighting scheme: the $\mu$-dependent terms do not depend on the source $s$, so for a fixed target word $t$ they do not affect which source is its best translation, and they vanish entirely when $\mu = \mathbf{0}$.

\subsection{M-Step}
Next, we will describe the M-step. Given an optimal matching $\matching^\star$ computed
in \cref{sec:e-step}, we search for a matrix $\Omega \in
\mathbb{R}^{d \times d}$. We additionally
enforce the constraint that $\Omega$ is a real orthogonal matrix,
i.e., $\Omega^{\top}\Omega = I$.  Previous work
\citep{Xing2015,Artetxe2017} found that the orthogonality constraint
leads to noticeable improvements.

Our M-step optimizes two objectives independently.
First, making use of the result in \cref{sec:after-algebra}, we minimize the following:
\begin{equation}
\begin{aligned}\label{eq:m-step1}
  -\log \ptheta(T_{\matching^\star} &\mid S_{\matching^\star}, \matching^\star) \\
  &= \frac{1}{2}||T_{\matching^\star} - \Omega \, S_{\matching^\star}||_F^2 + C
\end{aligned}
\end{equation}
with respect to $\Omega$ subject to $\Omega^\top \Omega = I$.
Note we may ignore the constant $C$ during the optimization.
Second, we minimize the objective 
\begin{equation}\label{eq:m-step2}
  -\log \prod_{i \in \utrg} \ptheta(t_i) = \frac{1}{2}\sum_{i \in \utrg} ||t_i - \mu||_2^2 + D
\end{equation}
with respect to the mean parameter $\mu$, which is simply
an average. Note, again, we may ignore the constant $D$ during optimization.

Optimizing \cref{eq:m-step1} with respect to $\Omega$ is known as the
orthogonal Procrustes problem \citep{schonemann1966generalized,gower2004procrustes} and has a closed-form
solution that exploits the singular value decomposition \citep{horn1990matrix}. Namely, we
compute $U\Sigma V^\top = T_{\matching} S_{\matching}^{\top}$.
Then, we directly
arrive at the optimum:
$\Omega^\star = UV^\top$. Optimizing
\cref{eq:m-step2} can also been done in closed form; the point which minimizes distance to the data points (thereby maximizing the log-probability) is the centroid: $\mu^\star =
\sfrac{1}{|\utrg|} \sum_{i \in \utrg} t_i$.

\begin{algorithm}[t]
  \begin{algorithmic}[1]
    \Repeat
    \State \textit{// Viterbi E-Step}
    \State $\matching^\star \gets \argmax_{\matching \in \matchings} \log \ptheta(\matching \mid S, T)$
    \State $\utrg^\star \gets \Vtrg \setminus \{ i : (i, j) \in \matching^\star\}$
    \State \textit{// M-Step}
    \State $U\Sigma V^\top \gets \texttt{SVD}\left(T_{\matching^\star}S^{\top}_{\matching^\star}\right)$
    \State $\Omega^\star \gets UV^{\top}$
    \State $\mu^\star \gets \sfrac{1}{|\utrg^\star|} \sum_{i \in \utrg^\star} t_i$
    \State $\vtheta \gets \left(\Omega^\star, \mu^\star \right)$
  \Until {converged}
  \end{algorithmic}
  \caption{{\small Viterbi EM for our latent-variable model}}
  \label{alg:viterbi}
\end{algorithm}

\section{A Reinterpretation of \citet{Artetxe2017} as a Latent-Variable Model}\label{sec:reinterp}
The self-training method of \citet{Artetxe2017}, our
strongest baseline in \cref{sec:experiments}, may also be
interpreted as a latent-variable model in the spirit of
our exposition in \cref{sec:model}. Indeed, we only need
to change the edge-set prior $p(\matching)$ to allow
for edge sets other than those that are matchings. Specifically,
a matching enforces a one-to-one alignment between
types in the respective lexicons. \citet{Artetxe2017}, on
the other hand, allow for one-to-many alignments. We show how
this corresponds to an alignment distribution that is equivalent
to IBM Model 1 \citep{Brown:1993:MSM:972470.972474}, and that \citet{Artetxe2017}'s self-training
method is actually a form of Viterbi EM.

To formalize \citet{Artetxe2017}'s contribution as a latent-variable
model, we lay down some more notation.  Let $\alignments = \{1, \ldots, \nsrc+1\}^{\ntrg}$, where
we define $(\nsrc+1)$ to be \texttt{none}, a distinguished
symbol indicating unalignment. The set $\alignments$ is to be interpreted
as the set of all one-to-many alignments $\alignment$ on the bipartite
vertex set $V = \Vtrg \cupdot \Vsrc$ such that $a_i = j$ means the
$i^\text{th}$ vertex in $\Vtrg$ is aligned to the $j^\text{th}$ vertex
in $\Vsrc$. Note that $a_i = (\nsrc+1) = \texttt{none}$ means that the
$i^\text{th}$ element of $\Vtrg$ is unaligned. Now, by analogy to our formulation in \cref{sec:model}, we define
% Allow this derivation to break across the column boundary (scoped locally)
\begingroup
\allowdisplaybreaks
\begin{subequations}
\begin{align}
  \ptheta(T &\mid S) \coloneqq \sum_{\alignment \in \alignments} \ptheta(T \mid S, \alignment) p(\alignment) \displaybreak[3] \\
              &= \sum_{\alignment \in \alignments} \prod_{i=1}^{\ntrg}  \ptheta(t_i \mid s_{a_i}, a_i) p(a_i)  \\
  &= \prod_{i=1}^{\ntrg} \sum_{{a_i}=1}^{\nsrc+1} \ptheta(t_i \mid s_{a_i}, a_i) p(a_i)
\end{align}
\end{subequations}
\endgroup
The algebra given above is an instance of the dynamic programming trick introduced by \citet{Brown:1993:MSM:972470.972474}, which reduces the number of terms in the expression from exponentially many to polynomially many. 
We take $p(\alignment)$ to be a (parameterless) uniform distribution over all alignments.

\paragraph{\citeposs{Artetxe2017} Viterbi E-Step.}
Now, we can perform the maximization necessary for a Viterbi E-step with dynamic programming. 
Specifically, the maximization problem over alignments decomposes additively, i.e.,
\begin{equation}
\begin{aligned}
 \underset{\alignment \in \alignments}{\max} &\,\log \ptheta(\alignment \mid S, T) \\
 &= \sum_{i=1}^{\ntrg} \!\!\!\!\!\!\!\!\!\!\! \underset{\hspace{.75cm} 1 \leq a_i \leq (\nsrc+1)}{\max} \!\!\!\!\!\!\!\!\!\!\!\!\!\! \log \ptheta(a_i \mid S, T)
 \end{aligned}
\end{equation}
thus, we can simply find $\alignment^\star$ component-wise:
\begin{equation}
 a_i^\star = \!\!\!\!\!\!\argmax_{1 \leq a_i \leq (\nsrc+1)} \!\!\!\!\! \log \ptheta(a_i \mid t_i, s_{a_i})
\end{equation}
Note that there is no longer a computational need to fall back on the Viterbi approximation to EM---we can also efficiently compute the expectations necessary for EM with dynamic programming.

\paragraph{\citeposs{Artetxe2017} M-Step.}
The M-step remains unchanged from the exposition in \cref{sec:model} with
the exception that we fit $\Omega$ given submatrices~$S_{\alignment}$ and~$T_{\alignment}$ formed from a one-to-many alignment $\alignment$,
rather than a bipartite matching $\matching$.

\paragraph{Why a Reinterpretation?}
The reinterpretation of \citet{Artetxe2017} as a probabilistic model
yields a clear analytical comparison between our method and theirs. 
We see the \emph{only}
difference between the two models is the constraint on the bilingual lexicon
that the model is allowed to induce, i.e., 
our model enforces a one-to-one alignment whereas \citet{Artetxe2017} does not.

\begin{table*}[t]
  \centering
  \begin{adjustbox}{width=2\columnwidth}
\begin{tabular}{l c c c c c c c c c c c c}
\toprule
& \multicolumn{4}{c}{English--Italian} & \multicolumn{4}{c}{English--German} & \multicolumn{4}{c}{English--Finnish}\\
& 5,000 & 25 & num & iden & 5,000 & 25 & num & iden & 5,000 & 25 & num & iden \\
\midrule
\citet{Mikolov2013e} & 34.93 & 0.00 & 0.00 & 1.87 & 35.00 & 0.00 & 0.07 & 19.20 & 25.91 & 0.00 & 0.00 & 7.02 \\
\citet{Xing2015} & 36.87 & 0.00 & 0.13 & 27.13 & 41.27 & 0.07 & 0.53 & 38.13 & 28.23 & 0.07 & 0.56 & 17.95\\
\citet{Zhang2016m} & 36.73 & 0.07 & 0.27 & 28.07 & 40.80 & 0.13 & 0.87 & 38.27 & 28.16 & 0.14 & 0.42 & 17.56 \\
\citet{Artetxe2016} & 39.27 & 0.07 & 0.40 & 31.07 & 41.87 & 0.13 & 0.73 & 41.53 & \textbf{30.62} & 0.21 & 0.77 & 22.61\\
\citet{Artetxe2017} & 39.67 & 37.27 & 39.40 & 39.97 & 40.87 & 39.60 & 40.27 & 40.67 & 28.72 & \textbf{28.16} & 26.47 & 27.88 \\
Ours (1:1) & 41.00 & 39.63 & 40.47 & 41.07 & \textbf{42.60} & \textbf{42.40} & \textbf{42.60} & \textbf{43.20} & 29.78 & 0.07 & 3.02 & \textbf{29.76} \\
% Ours (1:1, rank 20k) & 40.87 & 39.67 & 40.87 & & 41.33 & 41.73 & 40.93 & & 27.67 & 25.98 & 27.04 \\
% Ours (1:1, rank 30k) & 40.80 & 40.53 & 41.60 & 40.20 & 41.93 & 41.33 & 42.00 & & 28.16 & 24.23 & 27.74 & \\
Ours (1:1, freq. constr.) & \textbf{42.47} & \textbf{41.13} & \textbf{41.40} & \textbf{41.80} & 41.93 & \textbf{42.40} & 41.93 & 41.47 & 28.23 & 27.04 & \textbf{27.60} & 27.81 \\
% Ours (1:1, rank 50k) & 41.67 & 40.87 & & & 42.00 & 41.33 & & 42.53 & 28.93 & 25.14 & 27.60 \\
\bottomrule
\end{tabular}
\end{adjustbox}
\caption{Precision at 1 (P@1) scores for bilingual lexicon induction of different models with different seed dictionaries and languages on the full vocabulary.}
\label{tab:bdi-results}
\end{table*}

\begin{table}[t]
  \centering
    \begin{adjustbox}{width=\columnwidth}
\begin{tabular}{l c c c c}
\toprule
 &  & en-it & \multicolumn{2}{ c }{en-de} \\
 & Dict & WS & RG & WS \\ \midrule
% \citet{Luong2015b} & Europarl & .331 & .335 & .356 \\
\citet{Mikolov2013e} & 5k & .627 & .643 & .528 \\
\citet{Xing2015} & 5k & .614 & .700 & .595 \\
\citet{Zhang2016m} & 5k & .616 & .704 & .596 \\
\citet{Artetxe2016} & 5k & .617 & .716 & .597 \\ \midrule
\multirow{3}{*}{\citet{Artetxe2017}} & 5k& .624 & .742 & .616 \\
& 25 & .626 & .749 & .612 \\
& num & \textbf{.628} & .739 & .604 \\ \midrule
\multirow{3}{*}{Ours (1:1)} & 5k & .621 & .733 & \textbf{.618} \\
& 25 & .621 & .740 & .617 \\
& num & .624 & .743 & .617\\ \midrule
% \multirow{3}{*}{Ours (1:1, rank 30k)} & 5k & .624 & .737 & .611 \\
% & 25 & .619 & .745 & .606 \\
% & num & .623 & .749  & .607 \\ \midrule
\multirow{3}{*}{Ours (1:1, freq. constr.)} & 5k & .623 & .741 & .609 \\
& 25 & .622 & .753 & .609 \\
& num & .625 & \textbf{.755} & .611 \\
\bottomrule
\end{tabular}
\end{adjustbox}
\caption{Spearman correlations on English--Italian and English--German cross-lingual word similarity datasets.}
\label{tab:ws-results}
\end{table}

\section{Experiments}\label{sec:experiments}

We first conduct experiments on bilingual dictionary induction and
cross-lingual word similarity on three standard language pairs,
English--Italian, English--German, and English--Finnish.

\subsection{Experimental Details}

\paragraph{Datasets.}
For bilingual dictionary induction, we use the English--Italian dataset
from \citet{Dinu2015} and the English--German and English--Finnish
datasets by \citet{Artetxe2017}. For cross-lingual word similarity,
we use the RG-65 and WordSim-353 cross-lingual datasets for
English--German and the WordSim-353 cross-lingual dataset for
English--Italian by \citet{Camacho-Collados2015}.

\paragraph{Monolingual Representations.}
We follow \citet{Artetxe2017} and train monolingual representations with word2vec, CBOW, and negative sampling
\citep{Mikolov2013d} on a 2.8 billion word corpus for English (ukWaC
+ Wikipedia + BNC), a 1.6 billion word corpus for Italian (itWaC), a
0.9 billion word corpus for German (SdeWaC), and a 2.8 billion word
corpus for Finnish (Common Crawl).\looseness=-1

\paragraph{Seed Dictionaries.}
Following \citet{Artetxe2017}, we use dictionaries of 5,000 words, 25 words, and a numeral dictionary consisting of words matching the
\texttt{[0-9]+} regular expression in both vocabularies.\footnote{The resulting dictionaries contain 2772, 2148, and 2345 entries for English--\{Italian, German, Finnish\} respectively.} In line with \citet{Sogaard2018}, we additionally use a dictionary of identically spelled strings in both vocabularies.

\paragraph{Implementation.}
Similar to \citet{Artetxe2017}, we stop training when the
improvement on the average cosine similarity for the induced
dictionary is below \(1\times 10^{-6}\) between succeeding iterations. Unless stated
otherwise, we induce a dictionary of 200,000 source and 200,000 target
words as in previous work \citep{Mikolov2013e,Artetxe2016}. For optimal
one-to-one (1:1) alignment, we have observed the best results by keeping the top
$k=3$ most similar target words. 
If using a frequency constraint, we restrict the matching in the E-step to the top 40,000 words in both languages.\footnote{We validated both values with identical strings using the 5,000 word lexicon as validation set on English--Italian.} Finding the optimal alignment on the $200,000 \times 200,000$ graph takes about 25 minutes on CPU;\footnote{Training takes a similar amount of time to \citet{Artetxe2017} due to faster convergence.} with the frequency constraint, matching takes around three minutes.\looseness=-1

\paragraph{Baselines.}
We compare our approach with and without the frequency constraint to the original bilingual mapping approach by \citet{Mikolov2013e}. In addition, we compare with \citet{Zhang2016m} and \citet{Xing2015} who augment the former with an orthogonality constraint and normalization and an orthogonality constraint respectively. 
Finally, we compare with \citet{Artetxe2016}, who add dimension-wise mean centering to \citet{Xing2015}, and
to \citet{Artetxe2017}.
Both \citet{Mikolov2013e}
and \citet{Artetxe2017} are special cases of our framework and
comparisons to these approaches thus act as an ablation
study. Specifically, \citet{Mikolov2013e} does not employ orthogonal
Procrustes, but rather allows the learned matrix $\Omega$ to range
freely. Likewise, as discussed in
\cref{sec:reinterp}, \citet{Artetxe2017} make use of a
Viterbi EM style algorithm with a different prior over
edge sets.\footnote{We do not compare against other recent improvements, e.g., symmetric reweighting \citep{Artetxe2018}.}\looseness=-1

\subsection{Results}

We show results for bilingual dictionary induction in 
\cref{tab:bdi-results} and for cross-lingual word similarity in 
\cref{tab:ws-results}.  Our method with a 1:1 prior outperforms all baselines on English--German and English--Italian.\footnote{Note that results are not directly comparable to \citep{Conneau2018} due to the use of representations trained on different monolingual corpora (WaCKy vs. Wikipedia).} Interestingly, the 1:1 prior by itself fails on English--Finnish with a 25 word and numerals seed lexicon. 
We hypothesize that the bipartite matching prior imposes too strong of a constraint to find a good solution for a distant language pair from a poor initialization. With a better---but still weakly supervised---starting point using identical strings, our approach finds a good solution. Alternatively, we can mitigate this deficiency effectively using a frequency constraint, which allows our model to converge to good solutions even with a 25-word or numeral-based seed lexicon. 
The frequency constraint generally performs similarly or boosts performance, while being about $8$ times faster. 
All approaches do better with identical strings compared to numerals, indicating that the former may be generally suitable as a default weakly-supervised seed lexicon.
On cross-lingual word similarity, our approach yields the best performance on WordSim-353 and RG-65 for English--German and is only outperformed by \citet{Artetxe2017} on English--Italian WordSim-353.\looseness=-1

\section{Analysis} \label{sec:analysis}

\begin{figure}[h]
\centering
\includegraphics[width=\linewidth]{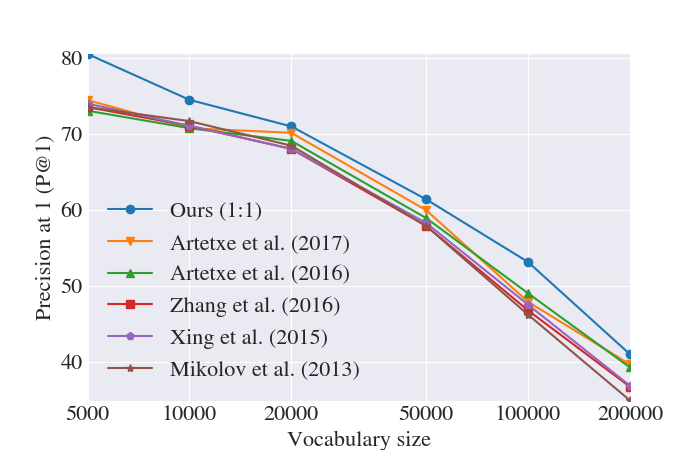}
\caption{Bilingual dictionary induction results of our method and baselines for English--Italian with a 5,000 word seed lexicon across different vocabulary sizes.}
\label{fig:vocab_sizes}
\end{figure}

\paragraph{Vocabulary Sizes.} 
The beneficial contribution of the frequency constraint demonstrates that in similar languages, many frequent words will have one-to-one matchings, while it may be harder to find direct matches for infrequent words. We would thus expect the latent variable model to perform better if we only learn dictionaries for the top $n$ most frequent words in both languages. We show results for our approach in comparison to the baselines in \cref{fig:vocab_sizes} for English--Italian using a 5,000 word seed lexicon across vocabularies consisting of different numbers $n$ of the most frequent words.\footnote{We only use the words in the 5,000 word seed lexicon that are contained in the $n$ most frequent words. We do not show results for the 25 word seed lexicon and numerals as they are not contained in the smallest $n$ of most frequent words.}
The comparison approaches mostly perform similarly, while our approach performs particularly well for the most frequent words in a language.

\begin{figure*}[!htb]
    \begin{subfigure}{.31\linewidth}
      \centering
         \includegraphics[height=1.4in]{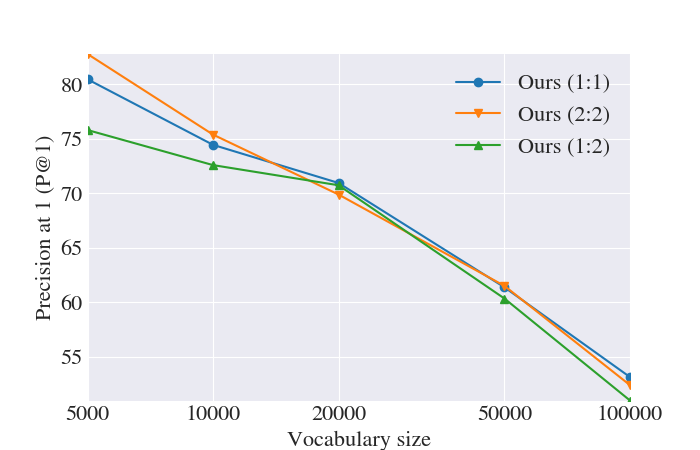}
         \caption{English--Italian}
    \end{subfigure}%
    \hspace*{0.3cm}
    \begin{subfigure}{.31\linewidth}
      \centering
         \includegraphics[height=1.4in]{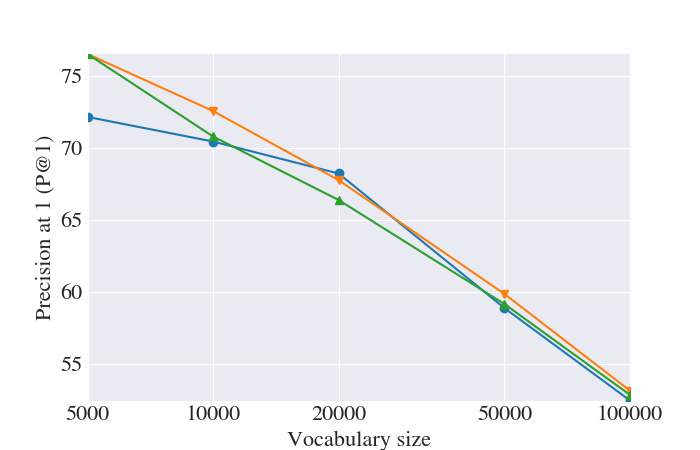}
         \caption{English--German}
    \end{subfigure}
    \hspace*{0.3cm}
    \begin{subfigure}{.31\linewidth}
      \centering
         \includegraphics[height=1.4in]{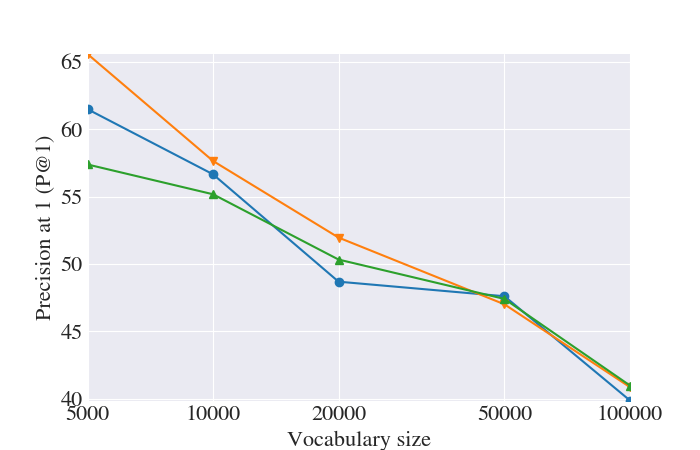}
         \caption{English--Finnish}
    \end{subfigure}
    \caption{Bilingual dictionary induction results of our method with different priors using a 5,000 word seed lexicon across different vocabulary sizes.}
\label{fig:priors}
\end{figure*}

\paragraph{Different Priors.} 
An advantage of having an explicit prior as part of the model is that we can experiment with priors that satisfy different assumptions. Besides the 1:1 prior, we experiment with a 2:2 prior and a 1:2 prior. For the 2:2 prior, we create copies of the source and target words $\Vsrc'$ and~$\Vtrg'$ and add these to our existing set of vertices $V' = (\Vtrg \cupdot \Vtrg',\Vsrc \cupdot \Vsrc')$. We run the Viterbi E-step on this new graph $G'$ and merge matched pairs of words and their copies in the end. Similarly, for the 1:2 prior, which allows one source word to be matched to two target words, we augment the vertices with a copy of the source words~$\Vsrc'$ and proceed as above. We show results for bilingual dictionary induction with different priors across different vocabulary sizes in  \cref{fig:priors}. 
The 2:2 prior performs best for small vocabulary sizes. As solving the linear assignment problem for larger vocabularies becomes progressively more challenging, the differences between the priors become obscured and their performance converges. 

\paragraph{Hubness Problem.} We analyze empirically whether the prior helps with the hubness problem. Following \citet{Lazaridou2015}, we define the \emph{hubness} $N_k(y)$ at $k$ of a target word $y$ as follows:
\begin{equation}
N_k(y) = |\{x \in Q \: | \: y \in \textit{NN}_k(x,G)\}| 
\end{equation}
where $Q$ is a set of query source language words and $\textit{NN}_k(x,G)$ denotes the $k$ nearest neighbors of~$x$ in the graph~$G$.\footnote{In other words, the hubness of a target word measures how often it occurs in the neighborhood of the query terms.} In accordance with \citet{Lazaridou2015}, we set $k=20$ and use the words in the evaluation dictionary as query terms. We show the target language words with the highest hubness using our method and \citet{Artetxe2017} for English--German with a 5,000 seed lexicon and the full vocabulary in \cref{tab:hubs}.\footnote{We verified that hubs are mostly consistent across runs and similar across language pairs.}
Hubs are fewer and occur less often with our method, demonstrating that the prior---to some extent---aids with resolving hubness. Interestingly, compared to \citet{Lazaridou2015}, hubs seem to occur less often and are more meaningful in current cross-lingual word representation models.\footnote{\citet{Lazaridou2015} observed mostly rare words with $N_{20}$ values of up to $50$ and many with $N_{20}> 20$.} For instance, the neighbors of \word{gleichg{\"u}ltigkeit} all relate to indifference and words appearing close to \word{luis} or \word{jorge} are Spanish names. 
This suggests that the prior might also be beneficial in other ways, e.g., by enforcing more reliable translation pairs for subsequent iterations.

\begin{table}[t]
  \centering
    %\begin{adjustbox}{width=\columnwidth}
\begin{tabular}{l l}
\toprule
\citet{Artetxe2017} & Ours (1:1) \\ \midrule
\multirow{2}{*}{\word{luis} (20)} & \word{gleichg{\"u}ltigkeit} \\
& - `\word{indifference}' (14) \\
\word{ungarischen}  & \word{heuchelei}  \\
- `\word{Hungarian}' (18) & - `\word{hypocrisy}' (13) \\
\word{jorge} (17) & \word{ahmed} (13) \\
\multirow{2}{*}{\word{mohammed} (17)} & \word{ideologie} \\
& - `\word{ideology}' (13) \\
\word{gewi{\ss}} & \multirow{2}{*}{\word{eduardo} (13)} \\
- `\word{certainly}' (17) & \\
\bottomrule
\end{tabular}
%\end{adjustbox}
\caption{Hubs in English--German cross-lingual representation space with degree of hubness. Non-name tokens are translated.}
\label{tab:hubs}
\end{table}

\paragraph{Low-resource Languages.}
Cross-lingual representations are particularly promising for low-resource languages, where few labeled examples are typically available, but are not adequately reflected in current benchmarks (besides the English--Finnish language pair). We perform experiments with our method with and without a frequency constraint and \citet{Artetxe2017} for three truly low-resource language pairs, English--\{Turkish, Bengali, Hindi\}. We additionally conduct an experiment for Estonian--Finnish, similarly to \citet{Sogaard2018}. For all languages, we use fastText representations \citep{Bojanowski2017} trained on Wikipedia, the evaluation dictionaries provided by \citet{Conneau2018}, and a seed lexicon based on identical strings to reflect a realistic use case. We note that English does not share scripts with Bengali and Hindi, making this even more challenging. We show results in \cref{tab:low-resource}. 
Surprisingly, the method by \citet{Artetxe2017} is unable to leverage the weak supervision and fails to converge to a good solution for English--Bengali and English--Hindi.\footnote{One possible explanation is that \citet{Artetxe2017} particularly rely on numerals, which are normalized in the fastText representations.} Our method without a frequency constraint significantly outperforms \citet{Artetxe2017}, while particularly for English--Hindi the frequency constraint dramatically boosts performance.
\begin{table}[t]
  \centering
     \begin{adjustbox}{width=\columnwidth}
\begin{tabular}{l c c c c}
\toprule
& en-tr & en-bn & en-hi & et-fi \\ \midrule
\citet{Artetxe2017} & 28.93 & 0.87 & 2.07 & 30.18 \\
Ours (1:1) & 38.73 & 2.33 & 10.47 & 33.79 \\
Ours (1:1, freq. constr.) & \textbf{42.40} & \textbf{11.93} & \textbf{31.80} & \textbf{34.78} \\
\bottomrule
\end{tabular}
 \end{adjustbox}
\caption{Results for English--\{Turkish, Bengali, Hindi\} and Estonian--Finnish.}
\label{tab:low-resource}
\end{table}

\paragraph{Error Analysis.} 

To illustrate the types of errors the model of \citet{Artetxe2017} and our method with a frequency constraint make, we query both of them with words from the test dictionary of \citet{Artetxe2017} in German and seek their nearest neighbors in the English representation space. P@1 over the German--English test set is 36.38 and 39.18 for \citet{Artetxe2017} and our method respectively. We show examples where nearest neighbors of the methods differ in \cref{tab:error-analysis}.
Similar to \citet{Kementchedjhieva2018}, we find that morphologically related words are often the source of mistakes. Other common sources of mistakes in this dataset are names that are translated to different names and nearly synonymous words being predicted. 
Both of these sources indicate that while the learned alignment is generally good, it is often not sufficiently precise.\looseness=-1

\section{Related Work}

\paragraph{Cross-lingual Representation Priors.} \citet{Haghighi2008} first proposed an EM algorithm for bilingual lexicon induction, representing words with orthographic and context features and using the Hungarian algorithm in the E-step to find an optimal one-to-one matching. 
\citet{Artetxe2017} proposed a similar self-learning method that uses word representations, with an implicit one-to-many alignment based on nearest-neighbor queries. \citet{Vuli2016a} proposed a more strict one-to-many alignment based on symmetric translation pairs, which is also used by \citet{Conneau2018}. Our method bridges the gap between early latent variable and word representation-based approaches and explicitly allows us to reason over its prior.

\begin{table}[t]
  \centering
     \begin{adjustbox}{width=\columnwidth}
\begin{tabular}{l l l l}
\toprule
Query & Gold & \citet{Artetxe2017} & Ours \\ \midrule
\word{unregierbar} & \word{ungovernable} & \word{untenable} & \word{uninhabitable} \\
% verschwendungssucht & wastefulness & extravagance & avarice \\
\word{nikolai} & \word{nikolaj} & \word{feodor} & \word{nikolai} \\
\word{memoranden} & \word{memorandums} & \word{communiqués} & \word{memos} \\
\word{argentinier} & \word{argentinians} & \word{brazilians} & \word{argentines} \\
\word{trostloser} & \word{bleaker} & \word{dreary} & \word{dark-coloured} \\
\word{umverteilungen} & \word{redistributions} & \word{inequities} & \word{reforms} \\
\word{modischen} & \word{modish} & \word{trend-setting} & \word{modish} \\
\word{tranquilizer} & \word{tranquillizers} & \word{clonidine} & \word{opiates} \\
\word{sammelsurium} & \word{hotchpotch} & \word{assortment} & \word{mishmash} \\
\word{demagogie} & \word{demagogy} & \word{opportunism} & \word{demagogy} \\
\word{andris} & \word{andris} & \word{rehn} & \word{viktor} \\
\word{dehnten} & \word{halmahera} & \word{overran} & \word{stretched} \\
\word{deregulieren} & \word{deregulate} & \word{deregulate} & \word{liberalise} \\
\word{eurokraten} & \word{eurocrats} & \word{bureaucrats} & \word{eurosceptics} \\
\word{holte} & \word{holte} & \word{threw} & \word{grabbed} \\
\word{reserviertheit} & \word{aloofness} & \word{disdain} & \word{antipathy} \\
\word{reaktiv} & \word{reactively} & \word{reacting} & \word{reactive} \\
\word{danuta} & \word{danuta} & \word{julie} & \word{monika} \\
\word{scharfblick} & \word{perspicacity} & \word{sagacity} & \word{astuteness} \\
\bottomrule
\end{tabular}
 \end{adjustbox}
\caption{Example translations for German--English.}
\label{tab:error-analysis}
\vspace{-5pt}
\end{table}

\section{Conclusion}
We have presented a novel latent-variable model for bilingual lexicon
induction, building on the work of \citet{Artetxe2017}. Our model combines the prior over bipartite
matchings inspired by \citet{Haghighi2008} and the discriminative,
rather than generative, approach inspired by
\citet{irvine-callisonburch:2013:NAACL-HLT}. We show empirical gains
on six language pairs and theoretically and empirically demonstrate the application of the bipartite matching prior to solving the hubness problem.

\section*{Acknowledgements}
The authors acknowledge Edouard Grave and Arya McCarthy, who provided feedback post-submission on the
ideas.  
Sebastian is supported by Irish Research Council Grant Number
EBPPG/2014/30 and Science Foundation Ireland Grant Number
SFI/12/RC/2289, co-funded by the European Regional Development
Fund. 
Ryan is supported by an NDSEG fellowship and a Facebook
fellowship. 
Finally, we would like to acknowledge Paula Czarnowska, who spotted a sign error in a late draft.\looseness=-1

\bibliographystyle{acl_natbib}
\bibliography{priors_bdi}

\onecolumn
\appendix

\section{Proof}\label{sec:theory}

\propAOne*
\begin{proof}
The result follows by expanding the two squared norms:
\begin{subequations}
\begin{align}
 \log \frac{\ptheta(t \mid s)}{\ptheta(t)}  &=  \log \ptheta(t \mid s)  - \log \ptheta(t) \\
&= -\sfrac{1}{2}||t - \Omega\, s||_2^2 + \sfrac{1}{2}||t - \mu||_2^2  \\
&= t^\top \Omega\, s - \sfrac{1}{2}||\Omega\, s||_2^2 {}- t^\top \mu + \sfrac{1}{2}||\mu||_2^2 \\
&= t^\top\!\left(\Omega\, s - \mu\right) - \sfrac{1}{2}||s||_2^2 + \sfrac{1}{2}||\mu||_2^2 \\
&= t^\top\!\left(\Omega\, s - \mu\right) + \sfrac{1}{2}\left(||\mu||_2^2 - 1\right)
\end{align}
\end{subequations}
The third line expands the norms and cancels $\sfrac{1}{2}||t||_2^2$; the fourth uses $||\Omega\, s||_2 = ||s||_2$ since $\Omega$ is orthogonal; and the last uses $||s||_2 = 1$. When $\mu = \mathbf{0}$, $t^\top \Omega\, s = \cos(t, \Omega\, s)$ because $||t||_2 = ||\Omega\, s||_2 = 1$, recovering $\cos(t, \Omega\, s) - \sfrac{1}{2}$.
\end{proof}

\propATwo*
\begin{proof}
This proof follows \citet{Haghighi2008}. 
First, consider the following manipulation
\begin{subequations}
\begin{align}
\matching^\star &= \argmax_{\matching \in \matchings} \log \ptheta(\matching \mid S, T) &  \label{eq:initial-opt} \\
&= \argmax_{\matching \in \matchings} \log \ptheta(T \mid S, \matching) + \log p(\matching) & \color{gray}{\text{(Bayes' rule)}}\\
&= \argmax_{\matching \in \matchings} \log \ptheta(T \mid S, \matching) & \color{gray}{\text{($p(\matching)$ is uniform)}}\\
 &= \argmax_{\matching \in \matchings} \log \!\!\!\prod_{(i, j) \in \matching}  \ptheta(t_i \mid s_j) \prod_{i \in \utrg} \ptheta(t_i) & \color{gray}{\text{(\Cref{eq:model})}}\\
  &= \argmax_{\matching \in \matchings} \sum_{(i, j) \in \matching} \!\!\! \log \ptheta(t_i \mid s_j) + \sum_{i \in \utrg} \log \ptheta(t_i) & \color{gray}{\text{(log rules)}} \\
  &= \argmax_{\matching \in \matchings} \sum_{(i, j) \in \matching} \!\!\! \left(\log \ptheta(t_i \mid s_j) - \log \ptheta(t_i) \right)+ \underbrace{\sum_{i \in \Vtrg} \log \ptheta(t_i)}_{\text{constant}} & \color{gray}{\text{(add 0)}} \\
 &= \argmax_{\matching \in \matchings} \sum_{(i, j) \in \matching} \!\!\! \log \ptheta(t_i \mid s_j) - \log \ptheta(t_i) \label{eq:final-opt}
\end{align}
\end{subequations}
Next, assign each edge $(\vtrg(i), \vsrc(j))$ the weight
\begin{equation}
    w_{ij} \coloneqq \log \ptheta(t_i \mid s_j) - \log \ptheta(t_i) 
\end{equation}
Note that \Cref{eq:final-opt} maximizes over \emph{partial} matchings---a target vertex left unmatched contributes only its term $\log\ptheta(t_i)$, which was absorbed into the constant above---whereas the Hungarian algorithm returns a \emph{perfect} matching on a balanced graph. To reconcile the two, we augment the graph with a private dummy partner for each of the $\nsrc + \ntrg$ real vertices and join every vertex to its dummy by a $0$-weight edge. On this balanced graph, a maximum-weight perfect matching assigns each real vertex either to a real partner, when profitable, or to its dummy, i.e., leaves it unmatched; since the dummy edges carry weight $0$, no edge with $w_{ij} < 0$ is ever selected, and the real edges of the perfect matching form exactly the optimal partial matching. This is \emph{not} equivalent to computing a perfect matching on the original graph and deleting the nonpositive edges afterward, which can be suboptimal.
By the Hungarian algorithm \citep{kuhn1955hungarian}, this balanced assignment over $\nsrc + \ntrg$ vertices per side---and hence the argmax in \Cref{eq:initial-opt}---is solved in $\bigO((\nsrc+\ntrg)^3)$ time.
\end{proof}

\end{document}